\documentclass[10pt,twocolumn,letterpaper]{article}

\usepackage[pagenumbers]{cvpr} 
\usepackage[numbers]{natbib}

\usepackage{cuted}

\definecolor{cvprblue}{rgb}{0.21,0.49,0.74}
\usepackage[pagebackref,breaklinks,colorlinks,allcolors=cvprblue]{hyperref}

\def\paperID{*****} 
\def\confName{CVPR}
\def\confYear{2026}

\title{EndoLive: Real-Time Style Transfer for Endoscopic Endonasal Skull Base Surgical Video}

\author{Hurt, Griffin\\
University of Pittsburgh\\
School of Computing and Information\\
{\tt\small griffhurt@pitt.edu}
\and
Brinkman, Calvin\\
University of Pittsburgh\\
School of Computing and Information\\
{\tt\small cbrinkman@pitt.edu}
}

\begin{document}
\maketitle

\begin{strip}
  \centering
  \includegraphics[width=\linewidth]{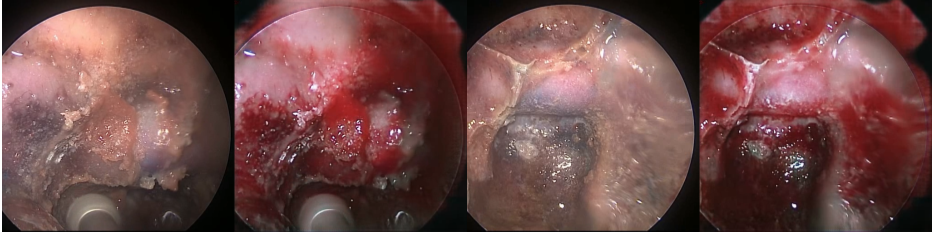}
  \captionof{figure}{\textbf{EndoLive in action.} Generated image pairs from the output of EndoLive. In each pair, the left image is a cadaveric image from our test dataset and the right image is the enhanced image produced in real-time by our model. The left pair of images demonstrates the capability of our model to generate fake blood in the visible field, a commonly described limitation of cadaveric models~\cite{supe2005cadaver}. The right pair of images demonstrates the capability of the model to semantically understand anatomical landmarks that appear similar in color in both live and cadaveric images. In this case, the pituitary gland is colored similarly between the original cadaveric frame and the generated image. }
  \label{fig:teaser}
\end{strip}

\begin{abstract}
Complex surgical procedures around critical anatomy, such as the endoscopic endonasal skull base surgery, requires significant practice and training on the part of the surgeon before they are allowed to perform the operation on a live patient. This training in typically done in cadaveric specimens, due to them containing the same critical structures as a living human. However, cadavers are not a perfect 1-to-1 substitute for a living patient. The dead and preserved tissues of a cadaver are colored completely differently than a living human, and -- without complex and expensive pumping systems -- do not bleed in the same way. As a result, identifying the critical pieces of anatomy that make this procedure so complex can be quite different in a live case than in a surgeon's cadaveric practice. This paper presents EndoLive, a framework for real-time style transfer between cadaveric endoscopic video and living human endoscopic video. Our method combines the ConStructS GAN model for realistic style transfer for surgical applications, with the HyPER-GAN model that can learn complex translations and perform them in real-time. We train EndoLive on unpaired cadaveric and live images taken from an endoscope, and test the trained model with cadaveric video, on a variety of devices. Experimental results demonstrate that EndoLive can perform cadaveric-to-live translation at speeds well above the minimum necessary for real-time, while maintaining semantic consistency of critical anatomical structures. Our source code is available at \texttt{https://github.com/griffhurt/endolive}.
\end{abstract}

\section{Introduction}
\label{sec:intro}
Endoscopic endonasal skull base (EESB) surgery is a novel form of neurological surgery that enables access to complex anatomy via an endoscope inserted into the nasal cavity~\cite{Ishii2014-xz}. This type of surgical procedure is used in a variety of neurological pathologies, including pituitary adenomas, craniopharyngiomas, and cerebral spinal fluid leaks (CSF leaks)~\cite{Ishii2014-xz}. Before the advent of skull base surgery, these diseases were treated with transcranial approaches that required long operative times~\cite{Ishii2014-xz}. However, in the late 1990s, two different independent research groups at University of Pittsburgh began pioneering a treatment mechanism that used novel endoscopic technology based on the prior work of ear, nose, and throat physicians~\cite{Maroon2005-wj}. Over the last 25 years, the technique has been further developed and is now a somewhat common practice in neurological surgery.

Due to the complexity of the procedure, surgeons wishing to specialize in this particular surgical style must spend a significant amount of time training for the approach. Some surgeons spend up to 9 years after medical school practicing the skull base approach before they are able to perform the procedure independently~\cite{Haynes2020-px}. The apprenticeship model for teaching new medical school graduates to become independent surgeons practicing the EESB approach results in a wide variance of mechanisms used in teaching and learning. James et al. reported that most hands-on training for the procedure occurs in an operative setting under the guidance of senior surgeons, an undeniably higher-risk approach compared to training in models~\cite{James2022}. This is in line with the classical ``Halstedian Preceptorship'' model used for teaching surgery, characterized with the aphorism ``see one, do one, teach one''~\cite{Kotsis2013}. However, when models are used for training EESB surgery, the broader literature suggests that cadaveric dissection models are the most common teaching tool employed~\cite{James2022}.

Cadaveric models are widely employed as a training simulation for surgical residency programs across a range of specialties, including spine surgery~\cite{Farah2023}, urology~\cite{AlJabir2020}, and orthopedic surgery~\cite{Wegmann2025}. While cadaveric models provide an accurate estimation of gross anatomy and allow surgeons to develop additional spatial reasoning skills, several limitations exist, including the absence of active bleeding, absence of breathing perception, and the malodorous nature of cadaveric models~\cite{supe2005cadaver}. Prior work has attempted to improve upon the realism of cadaveric models by perfusing blood through fresh cadavers~\cite{carey2015perfused}; however, this incurs a large cost in addition to the already high cost of cadaveric models.

Recent improvements in machine learning have opened the door for image generation models to create images from various other information modalities, including text and other images~\cite{elasri2022image}. Within the context of neural image generation, style transfer has become a common application. Style transfer is the process of changing the style of a given image without changing its underlying content~\cite{jing2019neural}. Style transfer has been used for a variety of purposes, both creative and technical, and executed with a variety of methods like diffusion models~\cite{zhang2023inversion} and GANs~\cite{ubhi2022neural,christophe2022neural}. Within the context of cadaveric models, style transfer can be used to eliminate limitations like poor coloration or lack of the appearance of blood by changing cadaveric video from an endoscope to appear like videos of live surgery. We call this process ``cadaveric to real style transfer" or ``CaRST''.

A nascent research area of EESB surgery that is gaining popularity is the integration of spatial computing technologies like augmented and mixed reality. Given the endoscopic nature of EESB, surgeons spend most of the case attending to several monitors throughout the operating room. Due to the sterility constraints of the operating room, surgeons are unable to adjust or otherwise manipulate the physical monitors surrounding the surgical field. This leads to communication overhead in the operating room for monitor layout, as the surgeon must ask other members of the surgical team to position their displays, and decreases overall surgical ergonomics. Spatial computing allows information to lose its physical boundaries and be presented on virtual elements that exist solely within the displays of a head-mounted display (HMD). Thus, the surgeon can manipulate these virtual displays without breaking sterility and gain a greater agency over their visual field and decrease split attention. Several papers have investigated the use of spatial computing in EESB~\cite{zeiger2020use, begagic2024augmented, li2016novel, choi2022application}, indicating its feasibility and benefits beyond traditional approaches. Using spatial computing to facilitate endoscopic surgery requires extremely low glass-to-glass latency~\cite{khan2022}, so integrating machine learning into these pipelines requires the models to have fast inference times and high frame rates to maintain standard of care.

In this work, we propose a dual-model approach for performing real-time endoscopic video style transfer from cadaveric based on ConStructS~\cite{ConstructS} and HyPER-GAN~\cite{HyperGAN}. While neither of these models on their own can effectively perform CaRST in real-time, combining them together enables novel surgical training visualization. We investigate the accuracy and efficiency of this architecture both on dedicated hardware (two NVIDIA RTX A5000s with CUDA) and on a spatial computing device (M2 Apple Vision Pro with CoreML) to understand its applicability to existing clinical workflow.

\section{Related Work}
\label{sec:related_work}
Style transfer is a popular and established topic of research in the field of computer vision due to its broad applications in a variety of areas. This project specifically investigates style transfer in the scope of real-time translation and surgical application. While we are among the first to look into this particular combination, each of these scopes has been researched independently by others, before us.
\subsection{Style transfer for surgery}
\subsubsection{Laparoscopic image-to-image translation}
In July of 2019, a group of researchers published a paper investigating the use of unpaired image-to-image translation for laparoscopic surgery \cite{Lapro}. The goal of the study was to take fake laparoscopic images rendered from a computer simulation, and to translate them to look like real images that could have been taken during a laparoscopic surgery. This synthetic data could then be used for a variety of tasks, namely image processing in deep-learning networks. The researchers' methods focused on extending the CycleGAN to the surgical application by invoking a type of loss that incorporates structural similarity between two images into its calculations. The paper resulted in a large set of synthetic laparoscopic images, with quality akin to real images taken during a surgery.
\subsubsection{Exploring semantic consistency}
In January of 2024, the same group of researchers published a follow-up to their previous research in generating synthetic laparoscopic images \cite{ConstructS}. This new paper focused entirely on how to preserve the semantic consistency of the fake input image, when translating it to a realistic style. The result of the paper is presenting the ConStructS (Contrastive Learning with Structural Similarity) model, developed specifically for the task of unpaired image-to-image translation while maintaining semantic consistency. Rather than using additional generators and a discriminator for unpaired translation like the CycleGAN, the ConStructS model uses an encoder-decoder method for image translation, then makes use of two types of loss to train the model. Patch loss compares the encoder outputs of the original image and the generated image while the semantic loss compares the structural similarity of both images, using the methods from the previous paper. The ConStructS model proves to be a simple and reliable method for unpaired one-way image-to-image translation, but as it was designed for the generation of training data for other, deeper models, the researchers did not develop it to run fast enough to be real-time. Therefore, ConStructS by itself is not suitable for our application.

\subsection{Real-time style transfer}
\subsubsection{Enhancing photorealism enhancement}
In June of 2021, a group of scientists published a paper titled "Enhancing Photorealism Enhancement" \cite{EPE}. This paper focused on the task of image-to-image style transfer in the scope of video games, with the end goal of translating modern video game graphics to be photorealistic. The paper was specifically geared towards photorealism in video game cities and streets, and was able to use a variety of datasets of city and street images to train their model. Their approach involved using metadata for each input image that was only available due to the images being pulled from a simulation. This information, including details such as the normal, depth, albedo, and segmentation maps of the images (referred to as the G-buffers), allowed this model to reach incredible levels of photorealism, while preserving all of the semantic data of the images. However, despite the motivations of the paper being in video games, this model could not perform this advanced translation in real-time. Additionally, this method isn't practical for uses outside of translation rooted in simulations, as the G-buffer information isn't easily extracted in real-life contexts such as cadaveric surgery.
\subsubsection{Patch-based translation}
Finally, in March of 2026, a pair of researchers published a paper presenting a way to perform high quality image-to-image translation in real-time \cite{HyperGAN}. Building upon the work of previous researchers, including "Enhancing Photorealism Enhancement," the paper presented a hybrid patch-based model (HyPER-GAN) that used a complicated network of real, synthetic, and translated images, cropped into patches, to train a generator fast enough to run in real-time. Notably, HyPER-GAN was designed such that it is trained on a set of images that have already been translated by another model. This makes it incredibly flexible, as it can be applied to nearly any pre-existing style transfer model and use case.

\section{Data}
\label{sec:data}
While there are many extant endoscopic datasets, they are typically scoped to live surgery, and do not include cadaveric images, therefore, for this study we opted to construct our own dataset. We extracted our data from endoscopic videos that were recorded as part of pilot tests for a previous study \cite{choi2022application} and our ongoing research in skull base surgery.

\begin{figure}
    \centering
    \includegraphics[width=0.8\linewidth]{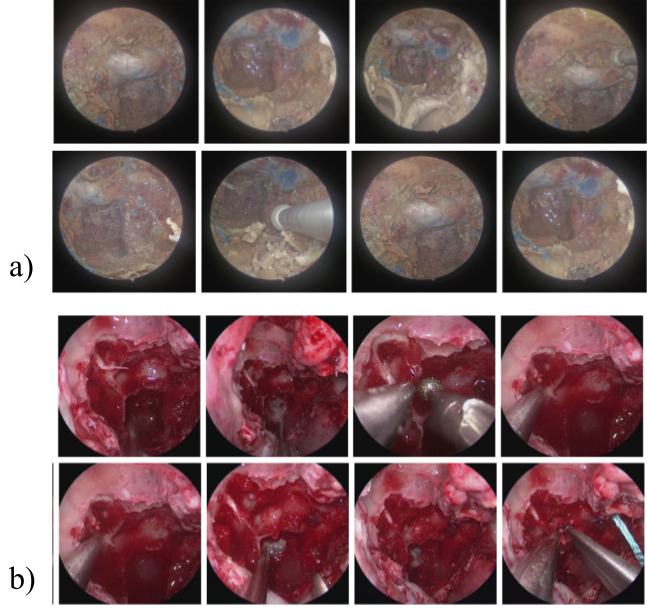}
    \caption{\textbf{Sample images.} a) Cadaveric and b) live images from the EndoLive dataset.}
    \label{fig:data}
\end{figure}

\subsection{Data Cleaning}
While EESB cases are generally mediated by the endoscope, there are several portions of the procedure that do not involve use of the endoscope, including closing (when the otolaryngology surgeon is suturing the patient's nose at the end of the procedure) and during tool changes. The raw videos from our dataset comprised the entire procedure, so these segments were originally included. We performed data cleaning by removing any segment of the video in which the endoscope was not within the nasal cavity. We removed these sections of video as the visual presentation of the patient's face is vastly different in cadaveric dissections compared to live surgery (in live surgery, the patient's face is entirely covered in surgical draping). Additionally, at these points in the surgery, the surgeon is no longer attending to the monitor and endoscopic feed, so the utility of the model during these segments is limited.

\subsection{Data Preprocessing}
Using an internal software tool, we extracted frames from 5.45 hours of live EESB surgery and from 4.76 hours of cadaveric EESB trials at 30 frames per second to create a dataset of 19,619 live and 17,151 cadaveric images of the nasal passage. Upon inspection, the extracted cadaveric images had a resolution of 1080 by 1080 pixels with the circular feed of the endoscope slightly inset within a black border, while the live images had a resolution of 720 by 720 pixels with the endoscope feed reaching edge of the images. To normalize the visual presentation of the two image sets, before training, cadaveric images were cropped to the bounding box of the circular feed, and all images were resized to 512 by 512 pixels to match the default input size of both models. Figure \ref{fig:data} depicts sample cadaveric and live images from the created dataset, before preprocessing.

\begin{figure*}[h]
  \centering
  \includegraphics[width=0.8\linewidth]{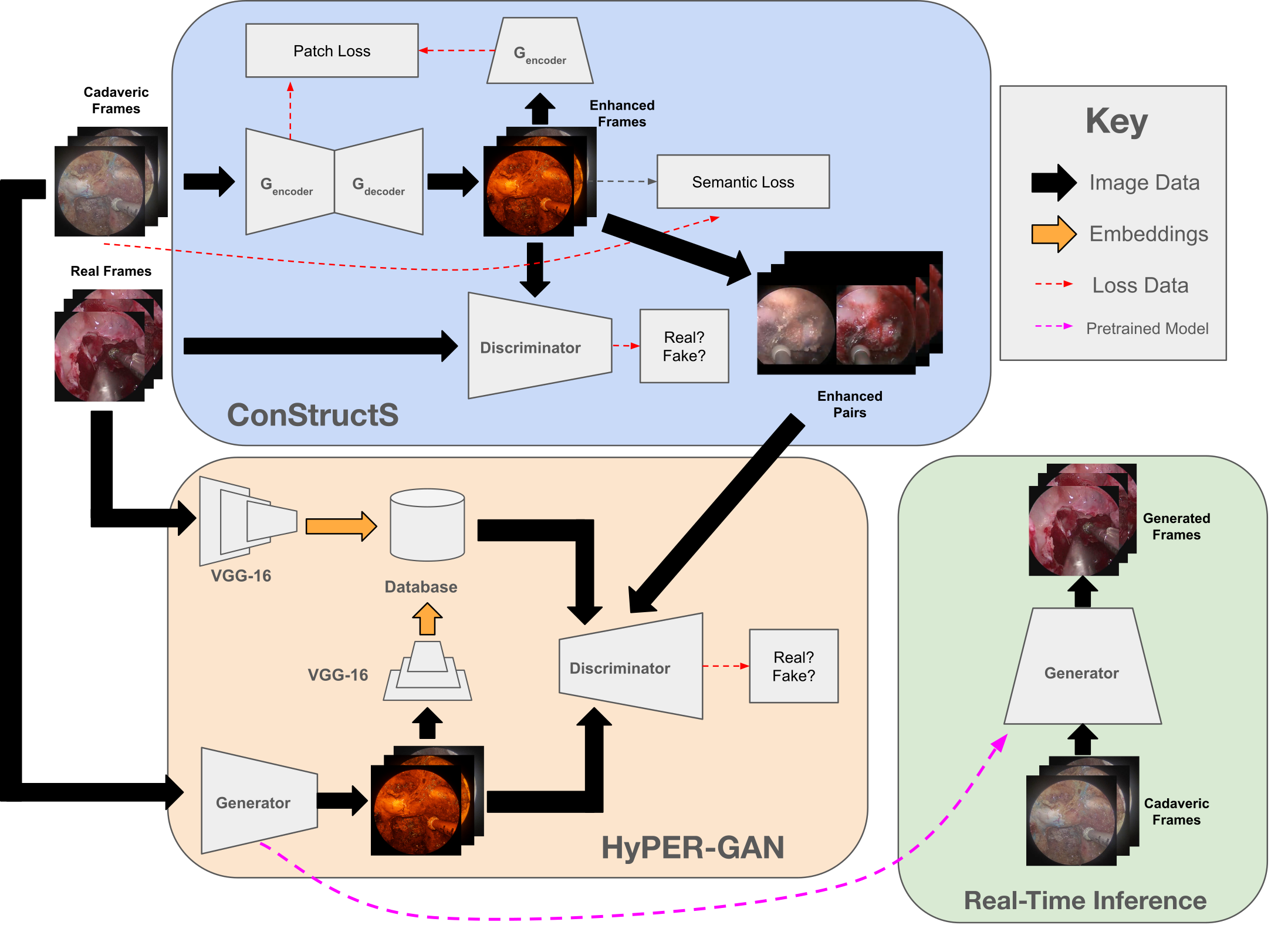}
  \caption{\textbf{EndoLive's architecture.} EndoLive uses a dual model architecture, expanding from ConStructS~\cite{ConstructS} and HyPER-GAN~\cite{HyperGAN} to perform endoscopic style transfer for EESB surgery in real time (around 60fps on an Apple Vision Pro device).}
  \label{fig:architecture}
\end{figure*}
\section{Methods}
\label{sec:methods}
The architecture of EndoLive comprises a combination of the ConStructS model described in Venkatesh et al.~\cite{ConstructS} and the HyPER-GAN described in Pasios and Nikolaidis's paper~\cite{HyperGAN}. We chose to use a dual-model approach, as neither model is suitable for real-time surgical style transfer on its own. Upon initial testing, the ConStructS ~\cite{ConstructS} model performed inference at roughly 10 frames per second on dual NVIDIA RTX A5000s, which indicated that, despite its high quality style transfer capabilities, it was unable to produce video output in real time on an embedded spatial computing headset. HyPER-GAN~\cite{HyperGAN}, on the other hand, was designed to expand upon a prior work, Enhancing Photorealistic Enhancement (EPE)~\cite{EPE}, which generated photorealistic images from video game footage, and allow for real-time inference. Like many CycleGAN architectures, EPE does not require paired real-simulated pairs of images; however, EPE does require additional semantic information about the simulated and real frames like segmentation masks and depth maps, collectively known as G-buffers. While this information is available in a video game due to its simulated nature, such semantic information is unavailable for cadaveric endoscopic video, which acts as our ``simulated'' video source in this example. The HyPER-GAN architecture effectively acts as a knowledge distillation of a larger EPE model with a smaller architecture to facilitate real-time inference and less GPU memory utilization. Although training an EPE model for this use case is infeasible, the HyPER-GAN architecture only requires simulated-real image pairs to train. Were we able to generate paired cadaveric-real image sets, we could still utilize the real-time benefits of HyPER-GAN without the requirements of G-buffers and human or game engine provided semantic understanding of the scene.

From these observations, we devised the EndoLive architecture, demonstrated in \autoref{fig:architecture}. A larger ConStructS model is trained on unpaired cadaveric and live frames from surgical videos to produce a generator model that can produce live-looking images from cadaveric frames. That generator is then used to perform an offline inference on the entire cadaveric training set to produce realistic live-cadaveric image pairs. These pairs are then used to train a HyPER-GAN model to distill the knowledge of the larger, slower ConStructS model. This results in a real-time style transfer model that requires no additional semantic understanding of the video content, simply unpaired frames as input.

\subsection{Modifications to the HyPER-GAN Code}
The HyPER-GAN architecture requires images to be cropped, recolored, and normalized before they can be processed as input to the generator. In the original source code published on their GitHub, the developers of the HyPER-GAN perform these transformations on the CPU instead of the GPU. Additionally, the speed of inference is bottlenecked by I/O overhead of the OpenCV library performing the reading and writing of video. To overcome these issues, we modified the HyPER-GAN code in two ways. First, we moved the read/write operations of OpenCV to separate threads in a producer-consumer model. This way, reading the next frame is not bottlenecked by writing of the prior frame. In a real-world deployment of this system, the video reading overhead (decompression of an H.264 video file) would be nonexistent, as the frames would be coming directly from a capture card, and the writing overhead can be solved with the multithreaded approach we provide in our code. Second, we move the initial image normalization and axis permutation to the GPU as opposed to the CPU. This allows for faster execution of the operations.

\subsection{Note on Temporal Stability}
It is important to note that neither of the underlying model architectures we utilized in the development of this system are temporally aware. That is, neither model contains a recurrent component that would allow the current state to have input from a prior frame. While the model was observed during testing to have overall good temporal stability (likely due to training on many similar frames in succession and the relatively similar presentation of cadaveric and live anatomy), we acknowledge that long surgical videos may experience temporal instability like anatomy changing color and generating blood appearing and disappearing. This is a known limitation of the model architecture, and remains an open area of future research.

\section{Experiments}
\label{sec:experiments}

\subsection{Testing Data}
We extracted 3 minutes (10,800 frames) of additional cadaveric video data to use as a test set for our analysis. While the videos were taken of the same anatomy of the training set, the frames extracted were entirely unseen to the model before inference.

\subsection{Quality evaluation}
We used Kernel Inception Distance (KID) \cite{KID} as our primary quality evaluation metric. KID is a way to quantitatively measure the similarity between images, and is a standard method to evaluate the performance of GANs. We calculate $d_o$ as the KID between the original cadaveric images and the live images. This gives us a control "gap" between the images. We then calculate $d_t$ as the distance between the translated cadaveric images and the live images. We can then find the percentage of the original gap that the model fills $C$ by using the formula

\[C = \frac{d_o-d_t}{d_o}\]

\begin{figure}
    \centering
    \includegraphics[width=1\linewidth]{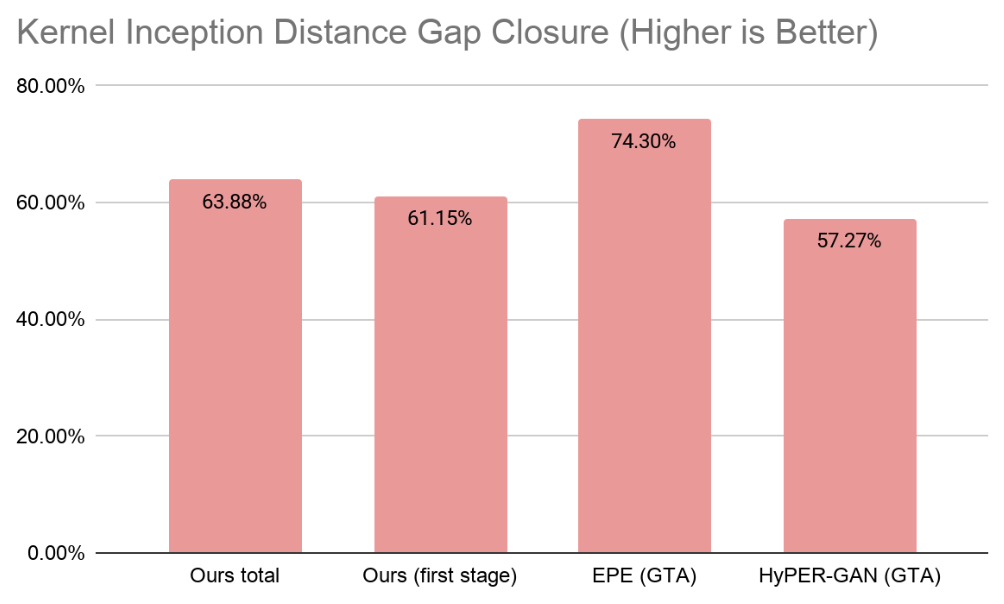}
    \caption{\textbf{Quality evaluation results} From left to right, the KID gap closure of EndoLive, EndoLive without HyPER-GAN implementation, Enhanced Photorealism Enhancement, and HyPER-GAN implementation of Enhanced Photorealism Enhancement.}
    \label{fig:kid}
\end{figure}

We use this "Gap Closure" method for evaluation, as it allows us to better compare the effectiveness of different model across different datasets. Figure \ref{fig:kid} displays the KID gap closure for EndoLive and Enhanced Photorealism Enhancement, each with and without HyPER-GAN implementation. Surprisingly, EndoLive with HyPER-GAN implementation actually outperforms the base ConStructS implementaion, and both outperform the HyPER-GAN implementation of Enhanced Photorealism Enhancement. The base EPE model without HyPER-GAN achieves the best results, which is unsurprising, of ~75\% gap closure, while EndoLive achieves a gap close of ~64\%.

Interestingly, the HyPER-GAN model trained on ConStructS was able to close more of the gap than the ConStructS model on its own. The reasons for this are speculative, but the HyPER-GAN model may be able to understand the semantic similarity between the live and cadaveric paired images via leveraging the pretrained VGG-16 model during training.

\subsection{Speed evaluation}

\begin{figure}
    \centering
    \includegraphics[width=1\linewidth]{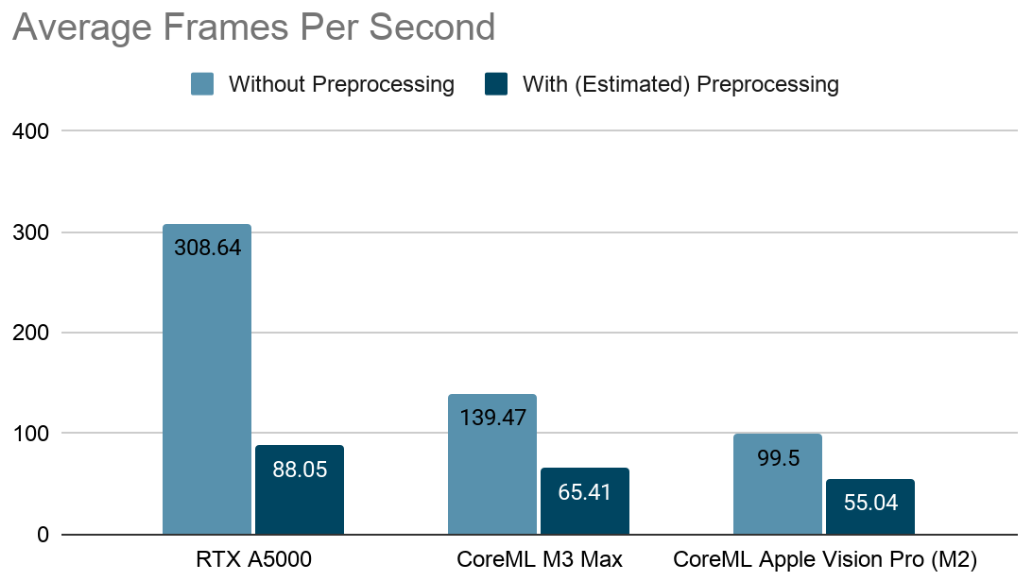}
    \caption{\textbf{Speed evaluation results} The average frames translated per second on the RTX A5000, M3 Max processor, and M2 processor; with and without preprocessing.}
    \label{fig:speed}
\end{figure}

We conducted three speed evaluations of our final model architecture: one with the machine we used to train the model (dual NVIDIA RTX A5000s connected in NVLink), one using the Apple Xcode CoreML profiler on a MacBook Pro with an M3 Max processor, and one using the CoreML profiler on the Apple Vision Pro with M2 processor. We used \texttt{coremltools}, a Python library developed by Apple to facilitate CoreML model development, and PyTorch's TorchScript JIT tracer (invoked via \texttt{torch.jit.trace}) to convert our final pretrained generator model to a CoreML bundle for deployment on Apple devices. We quantized the model to 16-bit floating point to facilitate faster inference on device. This bundle was then tested using the profiler tool built into Apple's Xcode software. Due to the model being trained on images of size 512x512, frame need to be preprocessed on each device before translation. Preprocessing the images within CoreML is not currently possible within the limitations of the architecture, so estimated preprocessing time is taken from the training machine preprocessing time. The results of this speed test are displayed in Figure \ref{fig:speed}.

Predictably, the dual RTX A5000 configuration performs the best out of the three, reaching an average of 308 frames translated per second if the frames are preprocessed ahead of time and 88 frames per second with preprocessing. The worst performing device was the Apple Vision Pro, but even while performing the preprocessing, it reaches a frame rate of 55 frames per second, well above the necessary 30 frames per second to keep up with a real-time video stream.

\section{Conclusion}
\label{sec:conclusion}
In this paper, we present EndoLive, a framework for performing real-time image-to-image translation of cadaveric endoscope video into living human endoscope video. We have shown that this framework is capable to achieving high quality synthetic endoscope images, comparable to established, high performant style transfer models. We have also demonstrated that it is capable of performing these translations in real-time; on a desktop computer, laptop, or even mixed-reality headset. This model has the potential to dramatically increase the realism of cadaveric trials for endoscopic endonasal skull base surgeries, reducing the learning curve of the procedure for new surgeons transitioning to living patients.
\subsection{Future Work}
A natural direction for future work is to evaluate EndoLive with a formal user study. It would be beneficial to expose real surgeons to the translated endoscope stream and evaluate whether EndoLive actually improves their ability to identify critical anatomy in live human patients and improves the training of the cadaveric trials. There are also plenty of other surgical procedures that use endoscopes or similar devices. EndoLive is flexible enough to be trained and applied to any of these use cases.

{
    \small
    \bibliographystyle{ieeenat_fullname}
    \bibliography{main}
}


\end{document}